\pdfoutput=1

\documentclass[11pt]{article}

\usepackage{acl}
\usepackage{nert}
\usepackage{times}
\usepackage{latexsym}
\usepackage[T1]{fontenc}
\usepackage[utf8]{inputenc}
\usepackage{microtype}

\newcommand*{\cat}[1]{{\textbf{\textsf{#1}}}}

\title{CuRIAM: Corpus re Interpretation and Metalanguage\\ in U.S. Supreme Court Opinions}

\usepackage{hyperref}

\author{Michael Kranzlein \\
  Georgetown University\\
  \eml{mmk119@georgetown.edu} \\\And
  Nathan Schneider \\
  Georgetown University \\
  \eml{nathan.schneider@georgetown.edu}\\ \AND
Kevin Tobia \\
  Georgetown University \\
  \eml{kevin.tobia@georgetown.edu}}

\begin{document}
\maketitle

\begin{abstract}
Most judicial decisions involve the interpretation of legal texts; as such, judicial opinion requires the use of language as a medium to comment on or draw attention to other language.
Language used this way is called metalanguage. We develop an annotation schema for categorizing types of legal metalanguage and apply our schema to a set of U.S.~Supreme Court opinions, yielding a corpus totaling 59k tokens. We remark on several patterns observed in the kinds of metalanguage used by the justices.
\end{abstract}

\section{Introduction}

U.S.~Supreme Court justices hear some of the most important cases in the country, resolving disagreements among lower courts, adjudicating the constitutionality of laws and regulations, and determining how those laws and regulations apply to real-world situations. Typically, a case might demand that the justices determine the meaning of just one word or phrase in a specific context. Two examples from the court's 2019 term are illustrative. In \href{https://www.supremecourt.gov/opinions/19pdf/17-1618_hfci.pdf}{\emph{Bostock~v.~Clayton County}}, the court interpreted the phrase ``because of... sex'' in the Civil Rights Act of 1964 as encompassing (and therefore making illegal) discrimination on the basis of \emph{sexual orientation}, a concept not directly identified in the language of the law. And in \href{https://www.supremecourt.gov/opinions/19pdf/18-801_o758.pdf}{\emph{Peter~v.~Nantkwest}}, the court held that even though a plaintiff must pay ``[a]ll the expenses of the proceedings'' in a challenge to an adverse decision by the Patent and Trademark Office (PTO), that phrase does not extend to cover the PTO's attorney's fees.

These and many other decisions rest on judgments about natural language: specifically, the meanings ascribed to legally binding text in statutes, regulations, and contracts as applied to a set of circumstances.
Moreover, judicial opinions are delivered in a natural language (namely, written English in the case of U.S. Supreme Court opinions).
They are therefore, to a large extent, metalinguistic: they feature language about language, or \textbf{metalanguage} \citep{berry2005}.

In the argumentation contained in the opinions, the justices \emph{quote} definitions from dictionaries; \emph{cite} precedents from prior rulings; \emph{apply rules} that have been established for legal interpretation; and \emph{present examples} showing terms could be used in ways that align with their interpretations. The purpose of this project is to quantify the frequency of these different types of metalanguage (among others) and analyze their use in judicial writing.

To this end, we introduce a schema that describes types of metalanguage in the legal domain and annotations of 32 U.S.~Supreme Court opinions from the 2019 term. Our contributions are the schema (\cref{sec:schema}), the annotated corpus (\cref{sec:data}), and a detailed analysis (\cref{sec:analysis}) revealing several patterns in how metalanguage is employed by the justices. We discuss relevant background for situating this study of metalanguage in \cref{sec:background} and its relationship to legal scholarship in \cref{sec:impact}.

This new resource is the first corpus of legal metalanguage and models an approach to annotation that may be adapted to other domains. Once released (pending expansion and revision based on the discoveries about the annotation scheme reported in this paper), it might be used as a tool for furthering legal and linguistic scholarship on judicial interpretation \citep{tobia2021a,gozdz-roszkowski2022} and help with the development of AI models of legal argumentation and reasoning \citep{atkinson2020,calegari2021}. It may also be useful for related NLP subtasks such as detecting citations and quotations.

\section{Background}\label{sec:background}
Definitions of metalanguage vary widely, but the metalanguage of interest for this paper is demonstrated well in \cref{ex:breyer}. In this example, Justice Breyer refers to a statute both as ``the Act'' and by its location in the U.S.~Statutes at Large, and he talks about a focal term in the case---``pollutant''---along with its definition.

\ex.\label{ex:breyer}First, the Act defines ``pollutant'' broadly, including in its definition, for example, any solid waste, incinerator residue, ``heat,'' ``discarded equipment,''' or sand (among many other things). §502(6), 86 Stat. 886.

Following \citet{berry2005}, we call the metalanguage in this example applied or \textbf{reflexive} metalanguage because it refers to the ``capacity of language to talk about itself'' \citep{sinclair1991}.%
\footnote{Berry distinguishes reflexive metalanguage from pure metalanguage and terminological metalanguage. Pure metalanguage comes from formal logic \citep{hilbert1928} and analytic philosophy \citep{tarski1933, quine1940} and involves the use of a metalanguage
to study an object language. This kind of metalanguage is often used in logic \citep{Williamson2014, dutta2016} and computer science \citep{chen2002, gluck2022}. Meanwhile, terminological metalanguage focuses narrowly on the vocabulary used to describe language. Several legal studies have adopted this latter definition of metalanguage, including one article proposing a neutral vocabulary for communicating about the law \citep{vaiciukaite2005} and two which take interest in a common set of concepts across national legal systems \citep{gunther2008, galdia2009}.}
The metalanguage we study in this paper is also \textbf{natural} because it does not involve artificial or formal languages.

In a series of papers in the early 2010s, Wilson brought a computational approach to natural metalanguage for the first time, and these works were pivotal for informing the creation of our schema. The first of these papers, \citet{wilson2010}, gave definitions of language mentions, metalanguage, and quotation, as well as an initial corpus of mentioned language. Then, \citet{wilson2011a}, \citet{wilson2011}, and \citet{wilson2012} iteratively built on this initial corpus, culminating in the \emph{enhanced cues} corpus, where stylistic cues (e.g.~quotation marks, italics, and bolding) and mention-significant words (e.g.~\w{meaning}, \w{name}, \w{phrase}) were used to identify candidate sentences that might contain metalanguage. The collection of mention-significant words was augmented using WordNet synsets, which helped expand the pool of candidate sentences. Any metalanguage in these candidate sentences was annotated and categorized according to a schema of four types. \Citet{wilson2013} presented the first automatic classifiers of natural metalanguage, and \citet{wilson2017} is a book chapter that provided an overview of metalanguage in NLP and noted the need for the development of new resources to aid the computational study of metalanguage. Since then, \citet{bogetic2021}---on metalanguage in Slovene, Croatian, and Serbian media articles and reader reactions---appears to be the only corpus of natural metalanguage published.

Other NLP research has explored the related topics of definitions, quotations, and citations. The Definition Extraction from Text (DEFT) corpus \citep{spala-19} was used in the 2020 SemEval shared task on definition extraction \citep{spala-20}. \Citet{hill-16} and \citet{yan-20} study the reverse dictionary task, where given a definition, the appropriate word has to be generated. And \citet{barba-21} propose a new task of exemplification modeling in which a word and its definition are provided and the expected output is a contextually appropriate example sentence using the word. There have also been many works studying quotation and citation: e.g., \citet{schneider-10} extract and visualize quotations from news articles; \citet{zhang2022} introduces a dataset for direct quote extraction; and \citet{carmichael2017} and \citet{lauscher2022} are two of many papers on legal and academic citation context analysis.

\begin{table*}[]
    \centering\small
    \begin{tabular}{|l|p{11cm}|}
        \hline
        \textbf{Category} & \textbf{Definition} \\
        \hline
        \cat{Focal Term} (FT) & Word or phrase whose meaning is under discussion in the case\\
        \hline
        \cat{Definition} (D) & Succinct, reasonably self-contained description of what a word or phrase means\\
        \hline
        \cat{Metalinguistic Cue} (MC) & Word or phrase cueing nearby metalanguage\\
        \hline
        \cat{Direct Quote} (DQ) & Span inside quotation marks that comes from an attributable source\\
        \hline
        \cat{Indirect Quote} (IQ) & Span inexactly recounting something that was said or written\\
        \hline
        \cat{Legal Source} (LeS) & Citation or mention appealing to a legal document or authority\\
        \hline
        \cat{Language Source} (LaS) & Citation or mention appealing to an authority on language\\
        \hline
        \cat{Named Interpretive Rule} (NIR) & Mention of a well-established interpretive rule or test used to support an argument about the meaning of a word or phrase\\
        \hline
        \cat{Example Use} (ES) & Intuitive, quoted, or hypothetical examples that demonstrate a word/term can or cannot be used in a certain way\\
        \hline
        \cat{Appeal to Meaning} (ATM) & A word or phrase indicating how one should go about interpreting the meaning other than by consulting an authoritative source or applying an interpretive rule\\
        \hline
    \end{tabular}
    \caption{Annotation Categories}
    \label{tab:cats}
\end{table*}

\section{Annotation Schema}\label{sec:schema}
Our annotation schema, which is the first to describe legal metalanguage, is given in \cref{tab:cats}. While the schema is technically flat, the ten categories can be thought of as falling into three broad groups: general metalanguage, quotes and sources, and interpretive rhetoric. The ten categories were refined through discussions among the authors and four law students who worked as annotators through several rounds of pilot annotation followed by a main annotation task.

\paragraph{General Metalanguage}
General metalanguage includes 3 categories in our schema: \cat{Focal Term}, \cat{Definition}, and \cat{Metalinguistic Cue}. Focal terms are words and phrases (often directly from statutes) that are central to the case. This includes terms that have nearby definitions, like \cref{ex:focal_term}, and those that don't, like \cref{ex:focal_term2}.\footnote{Examples given may not have all instances of metalanguage bracketed for the sake of readability and clarity related to the point each example supports.} An important characteristic of these terms is that they are mentions, not uses (see \citet{wilson2011}.

\ex.\label{ex:focal_term} The question presented: Does §924(e)(2)(A)(ii)'s ``[\textsubscript{FT} serious drug offense]'' definition call for a comparison to a generic offense?

\ex.\label{ex:focal_term2} I write separately to reiterate my view that we should explicitly abandon our ``[\textsubscript{FT} purposes and objectives]'' pre-emption jurisprudence.

Definitions are one of the most direct forms of metalanguage, as they are explicit statements that word $x$ means $y$. However, definitions proved nontrivial to bound. When they come from dictionaries, they are easy to identify, as in \cref{ex:def_violation}. There may also be formatting cues, which \cref{ex:def_causation} contains, that make definitions easy to spot.

\ex.\label{ex:def_violation} ...the term ``violation'' referred to [\textsubscript{D} the ``[a]ct or instance of violating, or state of being violated].'' Webster's New International Dictionary 2846 (2d ed. 1949) (Webster's Second).

\ex.\label{ex:def_causation} We have explained that ``[c]ausation in fact—i.e., [\textsubscript{D} proof that the defendant's conduct did in fact cause the plaintiff's injury] - is a standard requirement of any tort claim...''

But more complex examples led us to decide that definitions could also be more abstract \cref{ex:def_grievous}, non-comprehensive, or even negative \cref{ex:def_vehicle}---defining something by what it is not.

\ex.\label{ex:def_grievous} ...this Court has repeatedly explained that the rule of lenity applies only in cases of ```grievous''' ambiguity---[\textsubscript{D} where the court, even after applying all of the traditional tools of statutory interpretation, ```can make no more than a guess as to what Congress intended].'''

\ex.\label{ex:def_vehicle} ...the word ``vehicle,'' in its ordinary meaning, [\textsubscript{D} does not encompass baby strollers].

We also observed that the ``$x$ is $y$'' construction is not guaranteed to produce a definition, as in \cref{ex:baddef_disgorgement}, which offers a comment on the relevance of ``disgorgement'' without defining it.

\ex.\label{ex:baddef_disgorgement} Disgorgement is ``a relic of the heady days'' of courts inserting judicially created relief into statutes.

Metalinguistic cue is a category typically found near focal terms, definitions, and other types of metalanguage. These cues are single tokens like \emph{word}, \emph{means}, or \emph{phrase} that signal the author intends to talk about meaning. Other common instances are \emph{read}, \emph{interpret}, \emph{language}, \emph{terms}, and \emph{ambiguous}. Metalinguistic cues are not limited to single tokens \cref{ex:mc1}, and sometimes there can be many in a single sentence \cref{ex:mc2}:

\ex.\label{ex:mc1}First, ``based on age'' is an [\textsubscript{MC} adjectival phrase] that modifies the noun ``discrimination...''

\ex.\label{ex:mc2}In my view, however, the [\textsubscript{MC} provision] is also susceptible of the Government's [\textsubscript{MC} interpretation], i.e., that the entire [\textsubscript{MC} phrase] ``discrimination based on age'' [\textsubscript{MC} modifies] ``personnel actions.''

\citet{wilson2012} discusses stylistic cues as well as ``mention-significant words,'' which are similar to this category. We do not separately annotate stylistic cues like quotation marks and italics, but direct quote annotations do include the quotation marks.

\paragraph{Quotes and Sources}
This group consists of \cat{Direct Quote}, \cat{Indirect Quote}, \cat{Legal Source}, and \cat{Language Source}, which are fundamental to legal writing: ``The language of legal scholars and of advocates contains many quotations (laws, judgments, legal works) on which the author of the text comments. This is largely a matter of metalanguage'' \citep{mattila2006}.

Example \cref{ex:dq1} shows a common structure with a direct quote and its accompanying legal source.

\ex.\label{ex:dq1} An action under the [\textsubscript{LeS} FDCPA] may be brought [\textsubscript{DQ}``within one year from the date on which the violation occurs.''] [\textsubscript{LeS} §1692 k(d)]

In \cref{ex:las}, Justice Gorsuch refers to Black's Law Dictionary, one the most commonly cited language sources in Supreme Court opinions.

\ex.\label{ex:las} A principle is a ``fundamental truth or doctrine, as of law; a comprehensive rule or doctrine which furnishes a basis for others.'' [\textsubscript{LaS} Black's Law Dictionary 1417 (3d ed. 1933)]; [\textsubscript{LaS} Black's Law Dictionary 1357 (4th ed. 1951)]

The last category in this group is indirect quote. These spans are similar to direct quotes but are not verbatim and therefore not marked with quotation marks. We bound this category to paraphrasing and what could feasibly be uttered as part of a dialogue. This allows for constructions involving the verbs ``said'' and ``testified'' as shown in \cref{ex:iq} but usually excludes constructions with verbs such as ``claim,'' ``allege,'' and ``suggest.'' These latter verbs appear regularly in legal contexts but tend to be indicative of exposition or narration rather than dialogue. That is, these verbs are typically used to convey a legal position rather than to recount something that was previously spoken.

\ex.\label{ex:iq}But he testified in his deposition that [\textsubscript{IQ} he did not ``remember reviewing'' the above disclosures during his tenure].

\paragraph{Interpretive Rhetoric}
Finally, we have three of our most interesting metalanguage categories: \cat{Named Interpretive Rule}, \cat{Example Use}, and \cat{Appeal to Meaning}. Of these, named interpretive rules are the most straightforward. This category is intended to capture instances where justices invoke specific and established rules within the practice of law. Latin phrases like \cref{ex:noscitur} are common in this category, but other examples exist too, such as \cref{ex:categorical_approach}.

\ex.\label{ex:noscitur}...see id., at 21 (invoking the ``interpretive canon [\textsubscript{NIR} noscitur a sociis], a word is known by the company it keeps...''

\ex.\label{ex:categorical_approach} To determine whether an offender's prior convictions qualify for ACCA enhancement, we have used a ``[\textsubscript{NIR} categorical approach]...''

Example uses capture linguistic evidence, such as when justices quote statutes or famous works of literature to support a claim that a word can be used in a particular way:

\ex.\label{ex:equitable_principles} Congress itself has elsewhere used ``equitable principles'' in just this way: [\textsubscript{ES} An amendment to a different section of the Lanham Act lists ``laches, estoppel, and acquiescence'' as examples of ``equitable principles].''

Our last category is appeal to meaning, which covers the same kind of phenomenon as named interpretive rules, but in a broader sense. This category allows for general arguments, like \cref{ex:atm}, that suggest one linguistic interpretation is superior to another.

\ex.\label{ex:atm} We have stated in the past that [\textsubscript{ATM} we must ``read [the ADEA] the way Congress wrote it.'']

\section{Data Selection and Preprocessing}\label{sec:data}

For the development of CuRIAM, which stands for Corpus re Interpretation and Metalanguage\footnote{``Curiam'' and ``re'' are latin words commonly used in the legal profession meaning ``court'' and ``in the matter of / concerning,'' respectively.}, we retrieved opinion data from the 2019 term via the Harvard Caselaw Access Project\footnote{\url{https://case.law/}}. One of the authors who is a legal expert identified 18 cases that involved statutory interpretation (as opposed to, for example, exclusively procedural questions). From these 18 cases, we obtained 32 opinions\footnote{A Supreme Court case has more than one opinion when justices write concurring and/or dissenting opinions, in addition to the majority opinion.}.

As part of annotator\footnote{All four annotators were law students at the time of annotation with varying degrees of exposure to linguistics.} training, five opinions were annotated by all four annotators, and the results were discussed. Then, each of the 27 remaining opinions was randomly assigned to two annotators.

Preliminary quantitative analysis showed that in very long opinions the metalanguage often repeated, contributing less variety to the corpus. Therefore, in order to make the best use of our finite annotation budget, we truncated each opinion to around 2,000 tokens\footnote{For each opinion, to avoid awkward breaks in the text for annotators, we iteratively added the next paragraph until the number of tokens in the document exceeded 2,000. Therefore, some opinions have slightly more than 2,000 tokens}, prioritizing coverage of the Supreme Court term and the justices rather than coverage of individual, potentially very long opinions. The cutoff point of 2,000 tokens was motivated by a desire to include enough text to get past the narration and case summarization that is typical at the beginning of each opinion and into the core interpretive portion that tends to be rich in metalanguage.

Our truncation step affected 26 of our 32 opinions, meaning 6 opinions originally contained fewer than 2,000 tokens. All 6 of these were short concurrences. The median number of tokens per untruncated opinion was 4.5k, and the longest opinion contained almost 15k tokens. Annotating full opinions in the next version of the corpus is an obvious next step for the sake of completion.

We chose to start our study of legal metalanguage with U.S. Supreme Court opinions because they have broad impact and are well-known, but our schema could be applied to other types of legal documents
as well, particularly opinions from lower courts, where cases can still have significant impacts (e.g.~\href{https://ecf.flmd.uscourts.gov/cgi-bin/show_public_doc?2021-01693-53-8-cv}{\emph{Health Freedom Defense Fund~v.~Biden}}\footnote{In this case, a district court judge's interpretation of the term ``sanitation'' led to the nationwide transportation mask mandate being struck down.}) and contracts. An added benefit of studying Supreme Court opinions is that they feature high rates of metalanguage compared to some other legal documents\footnote{Preliminary explorations of the U.S.~Code of Federal Regulations, for example, revealed low rates of metalanguage, and the metalanguage that did appear was frequently limited to direct quotes from legal sources, featuring little interpretation or linguistic discussion of meaning.}
and more general language. \Citet{anderson2006} analyzed a sample of the British National Corpus \citep{bncconsortium2001} and found that only 11\% of sentences in their sample contained metalanguage. And of the metalanguage they identified, only 4\% was categorized as relating to ``language meaning.''

\begin{table}
    \centering\small
    \begin{tabular}{c cccc}
        \textbf{Justice} & \textbf{Maj} & \textbf{Conc} & \textbf{Diss} & \textbf{Total}\\ \midrule
        Alito & 3 & 1 & 2 & 6\\
        Breyer & 1 & 0 & 1 & 2\\
        Ginsburg & 2 & 1 & 0 & 3\\
        Gorsuch & 3 & 0 & 0 & 3\\
        Kagan & 2 & 0 & 0 & 2\\
        Kavanaugh & 2 & 2 & 1 & 5\\
        Roberts & 1 & 0 & 0 & 1\\
        Sotomayor & 2 & 3 & 0 & 5\\
        Thomas & 2 & 1 & 2 & 5\\ \midrule
        & 18 & 8 & 6 & 32\\
    \end{tabular}
    \caption{Opinions in corpus by justice: majority, concurring, dissenting}
    \label{tab:ops}
\end{table}

\Cref{tab:ops} shows the breakdown of the opinions we annotated by author and by opinion type. The corpus contains at least one majority opinion from each justice during the 2019 term, but 32 opinions is not enough data to make definitive claims about individual justices' metalanguage use or approach to interpretation. As such, most of our analysis focuses on observations about the schema itself and general patterns in the annotated data.

\section{Corpus Analysis}\label{sec:analysis}
The corpus contains 59,693 tokens with 6,088 annotated instances of metalanguage. Category frequencies and span lengths are given in \cref{tab:stats}. The two most common categories were direct quote and legal source, which accounted for two thirds of all annotations. On the other hand, several categories appeared fewer times than anticipated---we saw only 26 indirect quotes, 42 named interpretive rules, and 44 example uses. We noted considerable differences in the average length of annotated spans by category, and inter-annotator agreement varied, which is explored later in this section.

\begin{table}[]
    \centering\small
    \begin{tabular}{@{}l|c|c@{}}
    \hline
        \textbf{Category} & $n$ & \textbf{Mean Tok.~Len.~($\sigma$)} \\
        \hline
        Focal Term & \hphantom{0}778 & 3.9 (3.1)\\
        Definition & \hphantom{0}205 & 16.4 (10.8)\\
        Metalinguistic Cue & \hphantom{0}576 & 2.5 (2.2)\\
        Direct Quote & 1288 & 14.2 (14.9)\\
        Indirect Quote & \hphantom{00}26 & 19.0 (10.5)\\
        Legal Source & 2774 & 12.2 (13.0)\\
        Language Source & \hphantom{00}70 & 13.8 (7.7)\hphantom{0} \\
        Named Interpretive Rule & \hphantom{00}42 & 5.6 (5.3)\\
        Example Use & \hphantom{00}44 & 29.8 (22.5)\\
        Appeal To Meaning & \hphantom{0}165 & 10.1 (14.8)\\
    \end{tabular}
    \caption{Annotation category frequencies and span lengths}
    \label{tab:stats}
\end{table}

\paragraph{The most common types of metalanguage}
Direct quotes and legal sources are the most common categories of metalanguage in the opinions we analyzed---unsurprising since much of the argumentation the justices engage in revolves around the relation between the case at hand and relevant precedent. But it also seems that these are two of the easiest categories to annotate. Both categories had high inter-annotator agreement and are strongly signalled by formatting cues---usually quotation marks and parentheses. 

Example \cref{ex:sec} typifies a frequent pattern involving a direct quote and legal source, where a focal term of a case is introduced in quotation marks and relevant statutes are cited. This example also shows how categories of metalanguage are allowed to overlap. Focal terms were our third most common category and tended to include just a few tokens per span.

\ex.\label{ex:sec} ...the SEC may seek [\textsubscript{DQ} ``[\textsubscript{FT} disgorgement]''] in the first instance through its power to award [\textsubscript{DQ} ``[\textsubscript{FT} equitable relief]''] under [\textsubscript{LeS} 15 U. S. C. §78u(d)(5)]...

Metalinguistic cues were also frequent, and these spans had the fewest number of tokens. While easy to agree on when pointed out, in discussions, annotators commented that it was also easier to miss this type of metalanguage. Fortunately, the set of possible metalinguistic cues is relatively closed. Heuristic-based preannotations for this category and named interpretive rule could free up annotator time to focus on the categories where true disagreements arise more frequently.

\paragraph{The most challenging categories}
Annotation of the three interpretive rhetoric categories and indirect quotes yielded mixed results. These four categories were infrequent in our data, and annotators' conceptions of the categories varied. But two of these categories yielded useful results despite their relative rarity: named interpretive rule and appeal to meaning. In particular, we discovered that these two categories are challenging for opposing reasons. The former covers a more closed set of tokens: the names of well-established rules of legal interpretation. The latter is an open set, covering any span that suggests something about the meaning of a term. Despite being a more closed set, named interpretive rule suffered from low agreement among our annotators. We believe this to largely be a function of the category's rarity and a gap in training for the annotation task. On the other hand, appeal to meaning showed high agreement, especially when allowing partial matching. However, annotators still reported difficulty in annotation due to the broad definition of the category.

Example uses were difficult to identify in part because of their rarity, but also because of their diversity. Example uses can be quotes from statutes, references to prior cases, phrases from literary works, or sentences invented by a justice. And sometimes phrases which seem like they would cue an example use do not. In \cref{ex:eu}, ``for example'' is a clear indicator of an example use and a specific previous interpretation of a term is recounted:

\ex.\label{ex:eu} For example , [\textsubscript{EU} we have read the term ``equitable relief'' in the Employee Retirement Income Security Act of 1974 to refer to ``those categories of relief that were typically available in equity].''

But other times, no example use exists despite the presence of ``for example'':
\ex.\label{ex:eu2} The Act specifies, for example: that employers and employees must affirm in writing that the employee is authorized to work in the United States...

\paragraph{Agreement} Despite challenges with several categories that can be mitigated with additional annotator training and clarified guidelines, inter-annotator agreement results show the general validity of the schema. \Cref{tab:annotators} shows that our first three annotators, who joined the project at an earlier stage, had higher unlabeled agreement than our fourth annotator, who received the least training, suggesting that familiarity with the schema can help improve agreement. \Cref{tab:category_exact_match,tab:category_partial_match} show exact match and partial match F1, where each annotator's work is compared to all other annotations of the same documents the annotator was assigned. For example, suppose annotator 1 was assigned to documents A and B; annotator 2 was assigned to document A; and annotator 3 was assigned to document B. Annotator 1's precision and recall are calculated for each category, where the annotations of Annotator 2 and Annotator 3 are considered gold.

\begin{table}[]
    \centering\small
    \begin{tabular}{c|c|c|c}
        Annotator & P & R & F1\\
        A1 & 0.501 & 0.585 & 0.540\\
        A2 & 0.535 & 0.550 & 0.542\\
        A3 & 0.459 & 0.509 & 0.483\\
        A4 & 0.355 & 0.257 & 0.298\\
    \end{tabular}
    \caption{Unlabeled exact match F1 for each annotator. All other annotators' annotations considered gold while calculating annotator's F1.}
    \label{tab:annotators}
\end{table}

\begin{table*}
    \centering\small
    \begin{tabular}{c|c|c|c|c|c|c|c|c|c|c}
        Annot. & FT & D & MC & DQ & IQ & LaS & LeS & NIR & EU & ATM\\
        A1 & 0.391 & 0.240 & 0.347 & 0.620 & - &0.679 & 0.604 & 0.091 & 0.121 & 0.370\\
        A2 & 0.451 & 0.295 & 0.405 & 0.624 & - &0.824 & 0.577 & 0.174 & - &0.356\\
        A3 & 0.356 & 0.247 & 0.320 & 0.478 & - &0.769 & 0.547 & 0.091 & 0.146 & 0.432\\
        A4 & 0.298 & 0.057 & 0.190 & 0.090 & - &0.296 & 0.399 & - &0.118 & -\\
    \end{tabular}
    \caption{Category-based micro-average F1 for each annotator. All other annotators' annotations considered gold while calculating annotator's F1.}
    \label{tab:category_exact_match}
\end{table*}

\begin{table*}
    \centering\small
    \begin{tabular}{c|c|c|c|c|c|c|c|c|c|c}
    Annot. & FT & D & MC & DQ & IQ & LaS & LeS & NIR & EU & ATM\\
    A1 & 0.442 & 0.480 & 0.384 & 0.884 & - &0.943 & 0.956 & 0.182 & 0.424 & 0.731\\
    A2 & 0.508 & 0.564 & 0.431 & 0.912 & 0.111 & 1.000 & 0.986 & 0.348 & 0.054 & 0.693\\
    A3 & 0.446 & 0.466 & 0.336 & 0.927 & 0.095 & 0.923 & 0.906 & 0.182 & 0.293 & 0.724\\
    A4 & 0.396 & 0.426 & 0.243 & 0.881 & - &0.889 & 0.804 & - &0.235 & - \\
    \end{tabular}
    \caption{Category-based partial match micro-average F1 for each annotator. All other annotators' annotations considered gold while calculating annotator's F1.}
    \label{tab:category_partial_match}
\end{table*}

Exact match requires one annotator to mark the exact same span as the other annotator in order for it to be considered ``correct.'' Partial match allows for some flexibility, where if there is any overlap between two spans and they are marked with the same category, the annotation is considered correct. Partial matching provides insight when trying to understand how annotators approach longer spans and how features like punctuation impact agreement. For example, the last annotator to join the project (who had the least amount of training and time with the schema) did not realize that quotation marks should be included in direct quotes. As a result, we see a very low, outlier F1 score of .090 for direct quote in \cref{tab:category_exact_match}, but a score of .881 in \cref{tab:category_partial_match} with partial matching, which is in line with other annotators. We saw large increases from exact match to partial match in several other categories, including definition, language source, and legal source. These categories all had partial match F1 of .8 or higher. Indirect quotes were rare, and annotators expressed that challenging borderline cases were common among the candidate indirect quotes, leading to the lowest agreement out of all the categories, in both partial and exact match F1.

As noted earlier, metalinguistic cue, which saw only slight improvements from exact to partial match, seemed to show disagreement in part because metalinguistic cues are highly frequent and short in length, making them easy to miss. For example, one annotator marked ``interpretation'' as a metalinguistic cue in one sentence but not when it was used in a similar way three sentences later. This inconsistency was common, with only one annotator having high coverage of tokens they considered to be metalinguistic cues. Or see \cref{ex:mc3}, where the two annotators agreed on all spans except the metalinguistic cues and the token boundary of a definition. The second annotator did not mark ``word'' or ``read'' as metalinguistic cues, despite doing so in other documents.

\ex.\label{ex:mc3} In line with the rest of the [\textsubscript{MC} definition], the [\textsubscript{MC} word] [\textsubscript{DQ} ``[\textsubscript{FT} making]''] is most sensibly [\textsubscript{MC} read] to capture [\textsubscript{D}the entire process by which the contract is formed].

In addition to F1 measures of agreement, we calculated an average gamma score of .67 over the annotated documents. Gamma is a metric proposed in \citet{mathet2015a} that aims to address several complications in measuring agreement for annotations like ours, which have multiple labels, are span-based, and allow overlapping.

\section{Metalanguage and Law}\label{sec:impact}

Our corpus, CuRIAM, contributes to the study of reflexive metalanguage in legal writing. While metalanguage has been studied in other domains, it remains relatively unexplored in the legal domain. Only a couple of works consider the type of meaning-centric metalanguage we talk about in this paper (see \citet{plunkett2014,hutton2022}). Adjacent areas of study, like legal metadiscourse, rhetorical structure, and argumentation mining have received more attention \citep{tracy2020,yamada2019,mckeown2021,yamada2022}. \citeauthor{mckeown2021}'s corpus, while similar to ours in that it proposes a schema of metadiscourse in Supreme Court opinions, is different because it focuses on structure and author-audience interaction, rather than meaning.

While legal metalanguage has received less attention, it is highly relevant to modern legal theory and practice. Over the past few decades, ``textualism has come to dominate statutory interpretation'' in the United States \citep{krishnakumar2021}. Textualism directs interpreters to evaluate the ``ordinary meaning'' of statutes, and textualists rely on dictionary definitions, linguistic intuitions, and increasingly, corpus linguistics \citep{lee2017}.

Interpretation is essential to many other areas of law. For example, the interpretation of contractual language is the source of most contract litigation between businesses \citep{schwartz2009}, and  high-profile constitutional disputes often involve the interpretation of language in the constitution (see, for example, \href{https://www.supremecourt.gov/opinions/21pdf/19-1392_6j37.pdf}{\emph{Dobbs v.~Jackson Women's Health Organization}}).

As Hutton helpfully puts it, ``Judges are not professional linguists, but they are professional interpreters. Law has its own specialized and highly reflexive culture of interpretation, its own distinctive metalanguage, and an
open-ended set of rules, maxims, conventions, and practices'' \citep{hutton2022}. The systematic study of metalanguage in law can help uncover the nature of these interpretive practices. Not all of legal interpretive practice is obvious, as recent empirical studies have revealed \citep{krishnakumar2016}. Thus, discoveries about the practice of legal interpretation, via study of metalanguage, can provide important knowledge to legal practitioners, including judges themselves.

\section{Conclusion, Limitations, and Future Work}
This work describes an original schema for categorizing legal metalanguage and deploys it on U.S.~Supreme Court opinions, yielding a new corpus and an accompanying analysis. We commented on the frequencies of different types of legal metalanguage, and remarked on what went well in annotation, as well as several challenges. 

This work has several noteworthy limitations. First, this corpus contains data from only one Supreme Court term, authored by only 9 people. As such, it is not a representative sample of judicial language or even Supreme Court language, but rather a starting point for studying legal metalanguage. It also only covers English data from the U.S. judicial system. Second, the current version of the corpus is small and lacks adjudicated gold-standard labels; this data sparseness, rather than inherent ambiguity or subtleties in the language, may drive low accuracies in future classifiers. 

We are currently performing a round of revisions to address some of the issues with the guidelines, and to cover a larger number of opinions. Once complete, the corpus and annotation guidelines will be released publicly to encourage research on legal metalanguage, computational models thereof, and applications to legal interpretation.

\section*{Acknowledgements}
This research was supported in part by NSF award IIS-2144881 and in part by the Fritz Family Fellowship at Georgetown University. We are grateful to our annotators Marion Delaney, Tia Hockenberry, Danny Shokry, and Vito Whitmore, as well as to Lisa Singh for productive conversations about the schema. We also thank  those in our research lab who provided helpful feedback.

\bibliography{curiam,curiam_manual}

\bibliographystyle{acl_natbib}

\end{document}